\documentclass[letterpaper, 10 pt, conference]{ieeeconf}  
\IEEEoverridecommandlockouts                              
\usepackage{graphicx} 
\usepackage{epsfig} 
\usepackage{mathptmx} 
\usepackage{times} 
\usepackage{amsmath} 
\usepackage{amssymb}  
\usepackage[colorlinks=true, allcolors=blue]{hyperref}
\usepackage{multirow}
\usepackage{gensymb}
\usepackage[utf8x]{inputenc}
\usepackage{float}
\usepackage{subfigure}
\usepackage{cite}
\usepackage{stmaryrd}

\title{RaP-Net: A Region-wise and Point-wise Weighting Network to Extract Robust Features for Indoor Localization}

\author{Dongjiang Li$^{*2}$, Jinyu Miao$^{*3}$, Xuesong Shi$^{4}$, Yuxin Tian$^{3}$, Qiwei Long$^{2}$, Tianyu Cai$^{5}$,\\
Ping Guo$^{4}$, Hongfei Yu$^{1}$, Wei Yang$^{2}$, Haosong Yue$^{3}$, Qi Wei$^{1}$, Fei Qiao$^{1}$
\thanks{*Equal contributions.}
\thanks{$^{1}$Tsinghua University, Beijing, 100084 China.}
\thanks{$^{2}$Beijing Jiaotong University, Beijing, 100044 China.}
\thanks{$^{3}$Beihang University, Beijing, 100191 China.}
\thanks{$^{4}$Intel Labs China, Beijing, 100190 China.}
\thanks{$^{5}$Shanghai Jiao Tong University, Shanghai, 200240 China.}
\thanks{Corresponding authors: xuesong.shi@intel.com, qiaofei@tsinghua.edu.cn.}
}
\begin{document}
\maketitle

\begin{abstract}
Feature extraction plays an important role in visual localization. Unreliable features on dynamic objects or repetitive regions will interfere with feature matching and challenge indoor localization greatly. To address the problem, we propose a novel network, RaP-Net, to simultaneously predict region-wise invariability and point-wise reliability, and then extract features by considering both of them. We also introduce a new dataset, named OpenLORIS-Location, to train the proposed network. The dataset contains 1553 images from 93 indoor locations. Various appearance changes between images of the same location are included and can help the model to learn the invariability in typical indoor scenes. Experimental results show that the proposed RaP-Net trained with OpenLORIS-Location dataset achieves excellent performance in the feature matching task and significantly outperforms state-of-the-arts feature algorithms in indoor localization. The RaP-Net code and dataset are available at \url{https://github.com/ivipsourcecode/RaP-Net}.
\end{abstract}

\section{Introduction}
Visual localization aims to estimate the camera pose based on a given image taken in a mapped area. It is a fundamental problem for many artificial intelligence applications \cite{sattler2018benchmarking, middelberg2014scalable}. Indoor visual localization seems to be a more difficult problem due to the challenges coming from repetitive features, e.g. carpet textures and visually identical pillars, or dynamic objects, e.g. pedestrians and moved furniture \cite{taira2018inloc}.

Visual localization approaches can be categorized into two classes: feature matching-based ones and regression-based ones. Feature matching-based methods extract local features, which contain hundreds of keypoints and corresponding descriptors, and then match them across images or register with a 3D model to get an estimation of pose \cite{liu2017efficient}. The local features can also be aggregated into a global (image-wise) feature for efficient image retrieval from a large database \cite{yue2019robust}. Regression-based methods, on the other hand, train a machine learning model to directly regress coordinates \cite{brachmann2018learning} or poses \cite{kendall2015posenet} from images. They can be more efficient since only a single forward pass is needed for each query process, but the accuracy may not be as good as feature matching-based methods \cite{sattler2019understanding}.

In recent years, some works have proven that features based on deep convolutional neural network (CNN) can significantly outperform traditional hand-crafted ones in localization \cite{li2020dxslam}. However, existing CNN-based features are still insufficient to handle challenging indoor scenes since the models only estimate the point-wise reliability of each pixel, including repeatability and distinctiveness \cite{detone2018superpoint, dusmanu2019d2, yi2016lift, sarlin2019coarse}. The regions plays different importance in localization tasks. We should focus on invariable regions, not dynamic or repetitive areas, especially in indoor scenes. Extracting such information requires a large receptive field and prefers region-wise invariability. Attention mechanism, which derives from natural language processing \cite{attention}, can be customized to detect salient regions based on the high-level semantic information\cite{noh2017large, xin2019localizing}. However, training satisfied attention mechanism in indoor scenes still be a unsolved problem due to the lack of applicable training samples and network architectures. Our work aims to solve this issue.

In this paper, we propose a novel feature extraction method for visual localization in long-term and dynamic indoor scenes. The proposed method estimates not only the point-wise reliability but also the region-wise invariability, which refers to a probability of static and stable regions. Thus, the method is named Region-wise and Point-wise Weighting Network (RaP-Net). To train the proposed model, we introduce an indoor image dataset with typical appearance changes. The contributions of this work can be summarized as follows:

\begin{itemize}
\item We propose a novel CNN-based feature extraction method customized for indoor visual localization. It scores each pixel with two different weights. The region-wise weight reflects the invariability in the changing environment. The point-wise weight measures the reliability of each local feature.
\item We introduce a new dataset, named OpenLORIS-Location, to train the proposed network. The dataset contains various visual interference from the real world. To the best of our knowledge, it is the first indoor image dataset with location annotations for training attention mechanism.
\item Experiments on feature matching and visual localization tasks prove that the proposed method outperforms most of the state-of-the-art methods. The method and dataset are both open-sourced.
\end{itemize}

\section{Related Works}
In this section, we review existing works that are mostly related to contributions of this paper, namely, local features and indoor image datasets.

\subsection{Local Features} 
Since the appearance of SIFT \cite{lowe2004distinctive}, local features based on hand-crafted heuristics have been widely used in computer vision tasks, such as structure-from-motion (SfM) or simultaneous localization and mapping (SLAM) \cite{yu2018ds, li2020dxslam, campos2020orb}. After that, plenty of algorithms have been proposed for either approximating the image processing operators to gain computational efficiency or re-designing the detector and descriptor to achieve better performance. For example, SURF \cite{bay2006surf} uses Haar filters and integral images for fast keypoint detection and descriptor calculation. ORB \cite{rublee2011orb} applies oriented FAST detector and rotated BRIEF descriptor to speed up feature extraction and improve performance.

Learned local features have been recently developed to replace hand-crafted counterparts \cite{yi2016lift, noh2017large, detone2018superpoint, ono2018lf, sarlin2019coarse, dusmanu2019d2} and achieve better matching accuracy against illumination and viewpoint changes \cite{dai2019comparison, fischer2014descriptor}. Early works of learned local features introduce pipelines to imitate the traditional hand-crafted method that firstly detects keypoints in images and subsequently calculates descriptors. For example, in LIFT \cite{yi2016lift}, keypoints are detected and then cropped regions are fed to a second network to estimate the orientation before going through a third network to perform description. Later, the detection network of LF-Net \cite{ono2018lf} combines keypoint detection and orientation estimation, and outperforms LIFT. 

To improve the efficiency of training two independent detection and description networks, researchers have utilized a shared base net for both the detector and descriptor \cite{sarlin2019coarse, dusmanu2019d2, noh2017large}, which could significantly reduce the computational consumption. SuperPoint \cite{detone2018superpoint} presents a fully convolutional neural network, and it learns interest point locations and descriptors by self-supervision. HF-Net \cite{sarlin2019coarse} also adopts the shared encoder and proposes a coarse-to-fine method for large-scale localization. It jointly trains local features from SuperPoint decoder and global descriptors from NetVLAD decoder \cite{arandjelovic2016netvlad}. Dusmanu \textit{et al.} design D2-Net \cite{dusmanu2019d2} that declares the knowledge of keypoint detection and description can be further shared. We also follow this idea and use such a training strategy. As a comparison, we introduce more receptive information from multi-scale high-level feature fusion and pay close attention to the invariability of regions.

\begin{figure*}[t]
\centering
\includegraphics[width=0.95\textwidth]{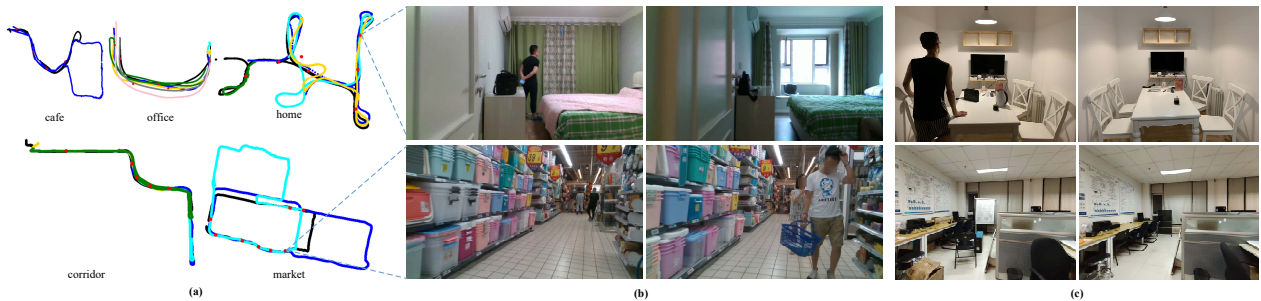}
\caption{Some samples from the OpenLORIS-Location dataset. (a)(b) Images of 30 locations are extracted from the OpenLORIS-Scene dataset. Red dots indicate the position of extracted locations and different sequences of the same scene are plotted in different colors. (c) Images of the other 63 locations are newly collected from real-world scenarios.}
\label{fig:dataset}
\end{figure*}

It should be noticed that not all local regions play equal importance in visual localization or SLAM system. It is important to extract and select keypoints located in static and distinct regions rather than dynamic or frequently changed objects. Some of the existing algorithms \cite{chen2017only, kim2017learned} are specially designed for visual place recognition, and others \cite{noh2017large, xin2019localizing} focus on a more general task, i.e. feature selection, which is more similar to our proposal. Such methods tend to detect stable or salient regions in the image where local features can be reliably extracted. Semantic segmentation \cite{paszke2016enet, ronneberger2015u, badrinarayanan2017segnet} can help to estimate the object-wise label, but unlike outdoor scenes, the dynamic and static attribute is hard to define in indoor environments. Thus, Noh et al. \cite{noh2017large} propose an attention-based network, named as DELF, to generate semantic local features for image retrieval. The model is trained on Google-Landmarks dataset by cross-entropy loss as an image classification problem. Xin et al. \cite{xin2019localizing} propose the landmark localization network to estimate the distinctiveness of each pixel and help to extract local features. Compared with these attention-based methods, our method works in a more difficult indoor environment with dynamic and daily changed elements, and we can directly extract the selected representative keypoints and descriptors.

\subsection{Visual Datasets for Training}
As mentioned before, attention mechanisms or other distinct region detection strategies can help localization algorithms hold their performances against dynamics and changed appearances. However, the related datasets for training such models are scarce and most of them are caught in outdoor environments. For instance, Retrieval-SfM dataset \cite{radenovic2016cnn}, used in \cite{xin2019localizing}, contains plenty of famous man-made architectures, such as palaces and towers, which are relatively static and positively contribute to visual place recognition. Google StreetView dataset, used in \cite{kim2017learned}, is another good source for obtaining images in urban scenes. These recent methods based on attention mechanism only work in the outdoor environment and they simply detect salient landmarks like buildings due to the training data. Thus, training attention to detect invariable regions in indoor scenes is still a bottleneck due to lacking training datasets similar to those outdoor ones.

The ideal dataset to train attention mechanism should contain acceptable viewpoint changes but large illumination changes and dynamic occlusions. Pictures describing the same location should be caught in different visual conditions. Unfortunately, existing indoor datasets can not meet these requirements. For example, InLoc dataset \cite{taira2018inloc} is introduced to SfM and re-localization. It contains typical dynamic disturbances but has severe viewpoint changes and almost unchanged illumination. Bicocca dataset \cite{bicocca} and EuRoC MAV dataset \cite{Burri25012016} are recorded in a relatively short period and contain less dynamic elements and illumination changes. They are excellent choices to evaluate performance but can not be used to train. OpenLORIS-Scene dataset \cite{shi2020we} contains consistent maps and collects several data sequences in each scene over a long period. Theoretically, it can be used for building our training dataset, but the number of locations may be a drawback and it can not be used to optimize networks as images are not split into distinct locations in the original paper. Thus, we construct our new dataset based on OpenLORIS-Scene and newly collected samples.

\section{The OpenLORIS-Location Dataset} 
To train the attention-based region-wise weight in our method, we introduce a new indoor image dataset, referred to as OpenLORIS-Location (LORIS stands for lifelong robotic vision as in \cite{shi2020we, she2020openloris}). The dataset is composed of 93 locations, and each location contains several images taken at different times with various kinds of typical indoor interference. Part of the data is derived from the OpenLORIS-Scene dataset \cite{shi2020we}, while others are newly collected in various indoor scenes. 

The OpenLORIS-Scene dataset is recently published and it provides real-world robotic data with more challenging factors, such as blur, texture-less images, dim lighting, and significant appearance changes. These visual elements can be used to train attention models to guide feature extraction for indoor localization. The dataset contains five typical indoor scenes, and ground-truth robotic poses in a consistent map are provided for all sequences. To construct our new dataset, we directly utilize the ground-truth pose from original paper \cite{shi2020we} and propose an incremental location extraction method. For each scene, we regard the position of the first RGB image from the first sequence as the first location. For the rest images in the scene, we impose the limitations of displacement and orientation to decide its location. Euclidean distances and changes of Euler angles are calculated between the position of the current image and all existing locations. If the closest distance and orientation are both less than the pre-defined threshold, the current image is added to the nearest location. To make sure images belonging to different locations have limited overlapped appearances, a new image is regarded as a new location only if it was far from existing locations or it had a large viewpoint change compared to existing locations. Fig. \ref{fig:dataset} gives us an illustration of extracted locations from each scene of OpenLORIS-Scene dataset and Fig. \ref{fig:dataset}(b) shows some samples in the home and market scene.

\begin{table}[t]
    \centering
    \caption{Statistical Description for Introduced New Dataset} 
    \begin{tabular}{c c c c}
    \hline
    Scene & Locations$^*$ & Images & Main challenge \\ 
    \hline
    office & 6+6 & 270 & illumination changes  \\
    home & 9+7 & 232 & moved furniture \\
    corridor & 12+5 & 291 & illumination changes, textureless \\
    market & 23+8 & 527 & high dynamics \\
    restaurant & 5+4 & 135 & high dynamics \\
    station & 8+0 & 98 & dynamics, repetitive pattern \\
    \hline
    Total & 63+30 & 1553 &  \\
    \hline
    \multicolumn{4}{l}{$^*$ (newly collected locations) + (derived from OpenLORIS-Scene) }\\
    \end{tabular}
    \label{tab:scene-number}
\end{table} 

On the other hand, to enlarge the scale of the dataset, we collect 63 new locations and add 1 typical scene, i.e. station. Those data are all taken by cameras on cell phones and most of them keep almost the same viewpoint, but they contain appearance changes and dynamic occlusions as shown in Fig.\ref{fig:dataset}(c).

Table \ref{tab:scene-number} summarizes the OpenLORIS-Location dataset. To the best of our knowledge, it is the first dataset of real-world indoor scenes that can be used for training attention mechanisms. In the 45 locations of office, home, and corridor, there are moderate scene changes and significant illumination changes. In the other 48 locations of market, restaurant, and station, illumination is relatively stable but there are numerous pedestrians and other dynamic objects. It can reflect the distribution of visual interference in real indoor scenes. In this paper, we use the part of this dataset to train the region-wise weight estimator of our RaP-Net. Although the dataset has relatively few images, we can mine numbers of triplets based on the location annotations, which meets the requirement of training.

Compared with the public indoor dataset \cite{taira2018inloc, bicocca, Burri25012016}, our new OpenLORIS-Location dataset has prominent advantages to train attention mechanism. The images of the same location in InLoc \cite{taira2018inloc} dataset sometimes have very few shared appearances due to large viewpoint changes, and they contain fewer dynamic occlusions. As for Bicocca \cite{bicocca} and EuRoC MAV \cite{Burri25012016}, visual interference is less included. Our dataset solves these drawbacks effectively and experimental results also conclude the effectiveness of the region-wise weight trained on the new dataset.

\begin{figure*}[t!]
\centering
\includegraphics[width=0.98\textwidth]{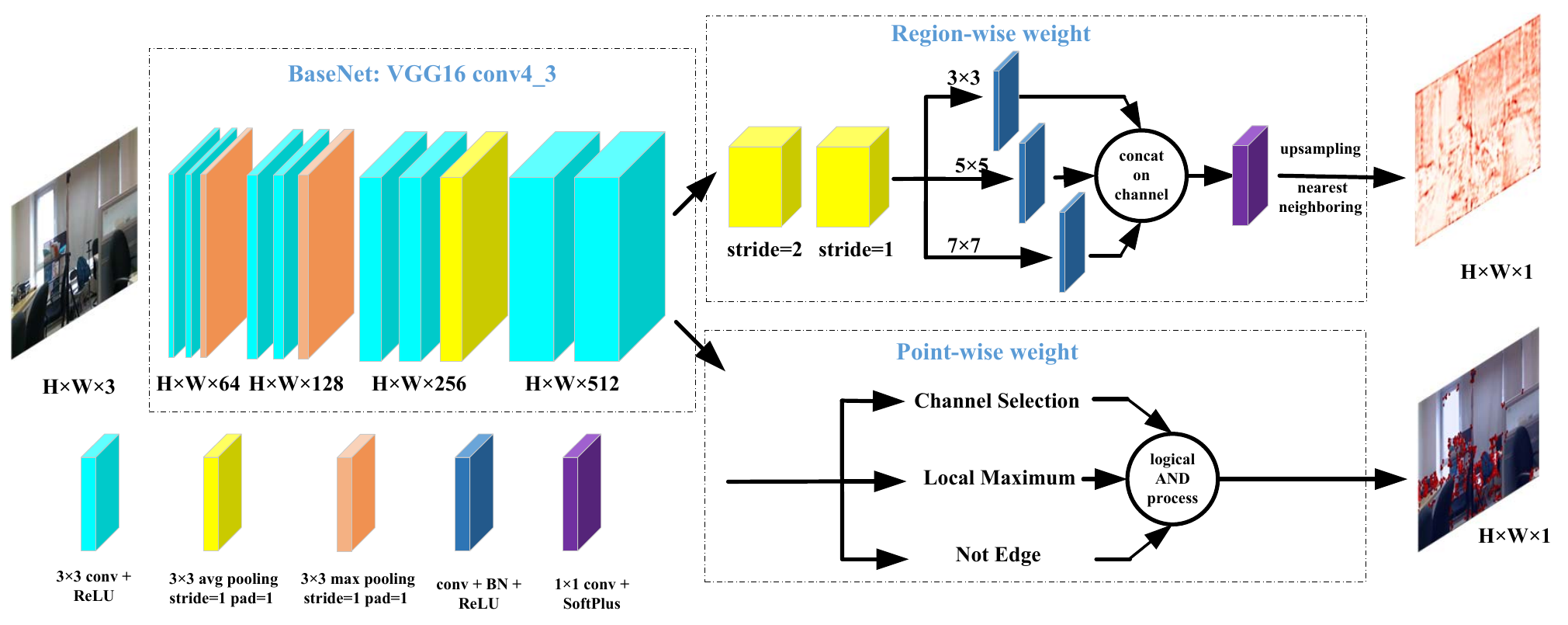}
\caption{Overview of our network. It consists of a shared backbone and two weight estimators that estimate point-wise reliability and region-wise invariability of local features.}
\label{fig:architecture}
\end{figure*}

\section{RaP-Net Methodology}
Our RaP-Net consists of a shared base net and two weight estimators as shown in Fig. \ref{fig:architecture}, the one calculates point-wise reliability of local features, and the other estimates a region-wise weight to select robust features. The region-wise weight is theoretically estimated by an attention mechanism, while the point-wise weight estimator is a feature detector. The dense feature maps provided by the base net are regarded as local descriptors. 

\subsection{Base Net: Feature Description}
In our paper, we adopt VGG-16 and truncate it after $\mathtt{conv4\_3}$ as our base net to extract dense feature maps. It is found that the reduced resolution of the feature map, where features are detected, may reduce the accuracy for strict feature matching. To get more accurate features while not increase the parameters of the network, we modify the stride of \emph{Max-Pooling} procedure to 1 in our base net so that we get a dense feature map having the same resolution with the input. Besides, we use $3\times 3$ kernels and add 1 zero-padding on both sizes in \emph{Max-Pooling} to increase the receptive field. The obtained dense feature map is denoted as $F \in \mathbb{R}^{512\times H \times W}$.

\subsection{Point-wise Weight: Reliability of Features}
Local features are detected by estimating the reliability of each pixel \cite{dusmanu2019d2}. For sake of convenience, we denote the $k^{th}$ channel of $F$ as $F^k$ and the feature vector at (i,j) as $F_{i,j}$. The detection strategy can be simply described as: A point (i, j) is detected only if $F^{k}_{i,j}$ is a local maximum in $F^{k}$ and $F^{k}_{i,j}$ has the maximal response in $F_{i,j}$. A detailed description of the detection method during the training and test process can be found in \cite{dusmanu2019d2}. We normalize the obtained point-wise weight map $P \in \mathbb{R}^{1\times H \times W}$ in image-level. A higher weight means the pixel has better reliability as a feature. 

\subsection{Region-wise Weight: Invariability of Regions}
To select features in static and distinct regions, we simultaneously apply an attention mechanism to estimate the invariability of regions. Motivated by \cite{xin2019localizing}, the architecture contains three parallel branches. As in Fig. \ref{fig:architecture}, each branch has a convolution layer with $3 \times 3, 5 \times 5, 7 \times 7$ kernel, respectively, for fusing features from various receptive fields. After batch-normalization (\emph{BN}) and being activated by \emph{ReLU} successively, the outputs of three branches are concatenated in channel dimension and fused by a $1 \times 1$ convolution. Activated by \emph{Soft-Plus} and image-level normalization, we can get a region-wise weight map $R \in \mathbb{R}^{1\times H \times W}$. It should be noticed that we use $2 \times 2$ \emph{Max-Pooling} to down-sample the dense feature $F$ at the beginning, and recover the resolution of the map by nearest neighboring (NN) interpolation. We find that such a procedure could boost the performance due to the larger receptive field of the module. The obtained region-wise weight indicates the invariability of a larger region, and it is used to re-weight the reliability of local features and help to select robust features.

\subsection{Training}
In our paper, we train the parameters of the base net and region-wise weight separately. The base net is pre-trained on ImageNet \cite{deng2009imagenet} and then fine-tuned on MegaDepth \cite{li2018megadepth} following the training strategy in \cite{dusmanu2019d2}. During training the region-wise weight, we freeze the parameters of the base net to not affect the reliability and apply metric learning to learn how to detect invariable regions. First, the dense feature map $F$ is seen as $H\times W$ local descriptors, and they are aggregated into a global descriptor $g$, which is a weighted sum of all descriptors from $F$.

\begin{eqnarray}
g = \sum_{i, j} R_{i,j} \times F_{i,j} 
\end{eqnarray}
where $i\in [1, H], j\in [1, W]$. After embedding, the global descriptor is normalized for metric learning.

\begin{figure*}[t]
\centering
\includegraphics[width=0.87\textwidth]{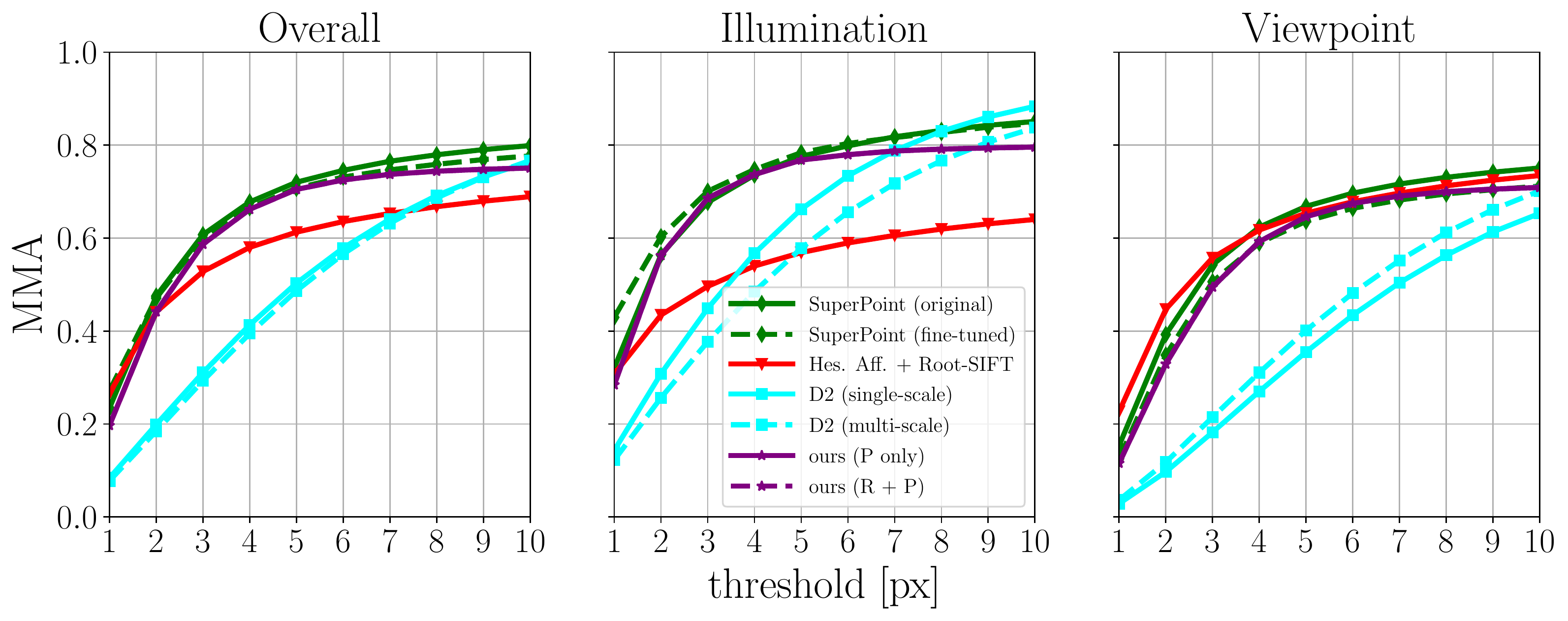}
\caption{Evaluation on HPatches. D2-Net features using a multi-scale process are extracted from three levels of image pyramids (0.5x, 1x, 2x).}
\label{fig:hseq}
\end{figure*}

We utilize our OpenLORIS-Location dataset as the training dataset. For each image, we regard it as an anchor and randomly selected images located in the same location as positives, images from different locations are seen as negatives. Then, we can obtain triplets containing 1 anchor image $a$, 1 positive $p$ and 4 negatives ${n}^{i}, i={1,2,3,4}$. During training,  we feed these triplets into our network and obtain the embedded global descriptors. In addition to the distance between anchor and positive, anchor and negatives, we also measure the mean distance between positive and negatives to apply in-triplet hard negative mining with anchor swap \cite{balntas2016learning}. Euclidean distance is used due to float descriptors.

\begin{eqnarray}
{d}_{a,n} = \frac{1}{4}\sum^{4}_{i=1} {||{g}_{a}, {g}^{i}_{n}||}_{2}, ~~
{d}_{p,n} = \frac{1}{4}\sum^{4}_{i=1} {||{g}_{p}, {g}^{i}_{n}||}_{2}
\end{eqnarray}
Where ${g}_{a}, {g}_{p}, {g}^{i}_{n}$ refers to the embedded descriptor of anchor, positive and $i^{th}$ negative, respectively. We swap $a, p$ when ${d}_{a,n} > {d}_{p,n}$, then $p$ becomes the anchor and $a$ is regarded as the positive sample in the current triplet. This procedure can make sure that the hardest negatives corresponding to the anchor are used for training in current step.

After that, we calculate ratio loss \cite{Elad2014deep} instead of commonly used triplet margin loss \cite{xin2019localizing, arandjelovic2016netvlad} to discard the influence of hand-crafted margin and boost the convergence of the network. The loss function is calculated as follows:

\begin{eqnarray}
L = {(\frac{{e}^{{d}_{a,p}}}{{e}^{{d}_{a,p}} + {e}^{{d}_{a,n}}})}^{2} 
+ {(1-\frac{{e}^{{d}_{a,n}}}{{e}^{{d}_{a,p}} + {e}^{{d}_{a,n}}})}^{2}
\end{eqnarray}
The ratio loss tend to make $\frac{{d}_{a,p}}{{d}_{a,n}} \longrightarrow 0$. In such a strategy, dynamics and frequently changed items in the images will be neglected to make ${d}_{a,p}$ become smaller, and meanwhile, invariable regions will be preferred. When ${d}_{a,n}$ gets larger, repetitive patches will be discarded and distinct textures can be emphasized.

\subsection{Inference} 
During inference, an input image is fed into the network and we get two weight maps $R$, $P$. $R$ provides a region-level invariability, higher weight means the point is located in an invariable region with higher confidence. $P$ provides pixel-level reliability of features, it can be used as a criterion to extract the local feature. We intend to detect reliable features in invariable regions and discard features in dynamic and varying regions. Thus, we use $R$ as a mask map to encourage the $P$ of invariable points and restrain the dynamic or repetitive points.

The average value of $R$ is calculated as an adaptive threshold $\overline{R}$. Points $(i,j)$ with $R_{i,j}$ higher than the threshold are encouraged, otherwise, they are restrained. Then we use $R$ to re-weight the reliability of features $P$ as follows:

\begin{eqnarray}
    score(i,j) = P_{i,j} \cdot {e}^{R_{i,j} - \overline{R}}
\end{eqnarray}

It can be seen that the $P_{i,j}$ of point with $R_{i,j}$ higher than the average will be multiplied by weight larger than 1, and the others will be suppressed with a small weight. Thus, a point with a higher score indicates that it is a reliable feature located in the invariable region, which is necessary for localization and tracking in indoor SLAM.

Here, instead of directly using $R$ to detect features \cite{xin2019localizing, noh2017large} or multiplying two weights, we re-weight point-wise reliability by region-wise invariability. It is because that $R$ is a region-wise invariability estimation and cannot reflect reliability of pixels to select accurate features, while $P$ cannot judge whether the feature is located in a dynamic region. Our proposed weighting strategy can successfully combine the advantages of two weights and benefit feature extraction.

\section{Evaluation}
In this section, we present the evaluation results of the proposed method on feature matching and indoor localization benchmarks. 

\subsection{Local Feature Matching}
From the aspect of local features, the proposed method is evaluated on the HPatches \cite{balntas2017hpatches} feature matching benchmark. Due to the limited resolution of frames used in practical applications, the maximal edge of images is resized below 640 pixels before extracting features, and then the keypoint positions are recovered to the original resolution for evaluation using the given ground-truth. We use mean matching accuracy (MMA) as a measurement under different matching thresholds following \cite{dusmanu2019d2, detone2018superpoint}, the results are shown in Fig.\ref{fig:hseq}. The accuracy under stricter threshold should be more concerned.

We separately evaluate our method with only consideration of $P$, named ``ours (P only)'', and the overall method using $R$ to re-weight $P$, named ``ours (R + P)''. It can be seen that our features using $R$ to re-weight $P$ have no further improvements due to very few prominent dynamic interference in HPatches dataset. Additionally, we finetune the original weight of SuperPoint\footnote{Original weight is from the official inference codes at \url{https://github.com/magicleap/SuperPointPretrainedNetwork}}, denoted as ``SuperPoint (original)'', on a hybrid training dataset (OpenLORIS-Location and MS-COCO 2014 \cite{coco})\footnote{We use unofficial codes from \url{https://github.com/eric-yyjau/pytorch-superpoint} since the authors do not provide related training codes.}. We cannot finetune D2-Net since the new dataset does not contain depth information and dense correspondences between pixels cannot be obtained. After fine-tuning, the matching accuracy of SuperPoint is slightly improved against illuminations, and it achieves a comparable overall performance with the original weight. It can be seen that extra training samples of indoor scenes only slightly improve the local features if we do not regulate the network properly. 

According to the figure, our proposed features outperform detect-and-describe features \cite{dusmanu2019d2} and traditional hand-crafted features for stricter matching threshold and achieve comparable performances with detect-then-describe features \cite{detone2018superpoint}. Compared with the baseline \cite{dusmanu2019d2}, the improvements are mainly caused by a feature map with undiminished resolution in the base net. It also concluded that the proposed region-wise weight does not affect the accuracy of the feature detector based on point-wise weight.

\begin{figure}[t]
\centering
\includegraphics[width=0.47\textwidth]{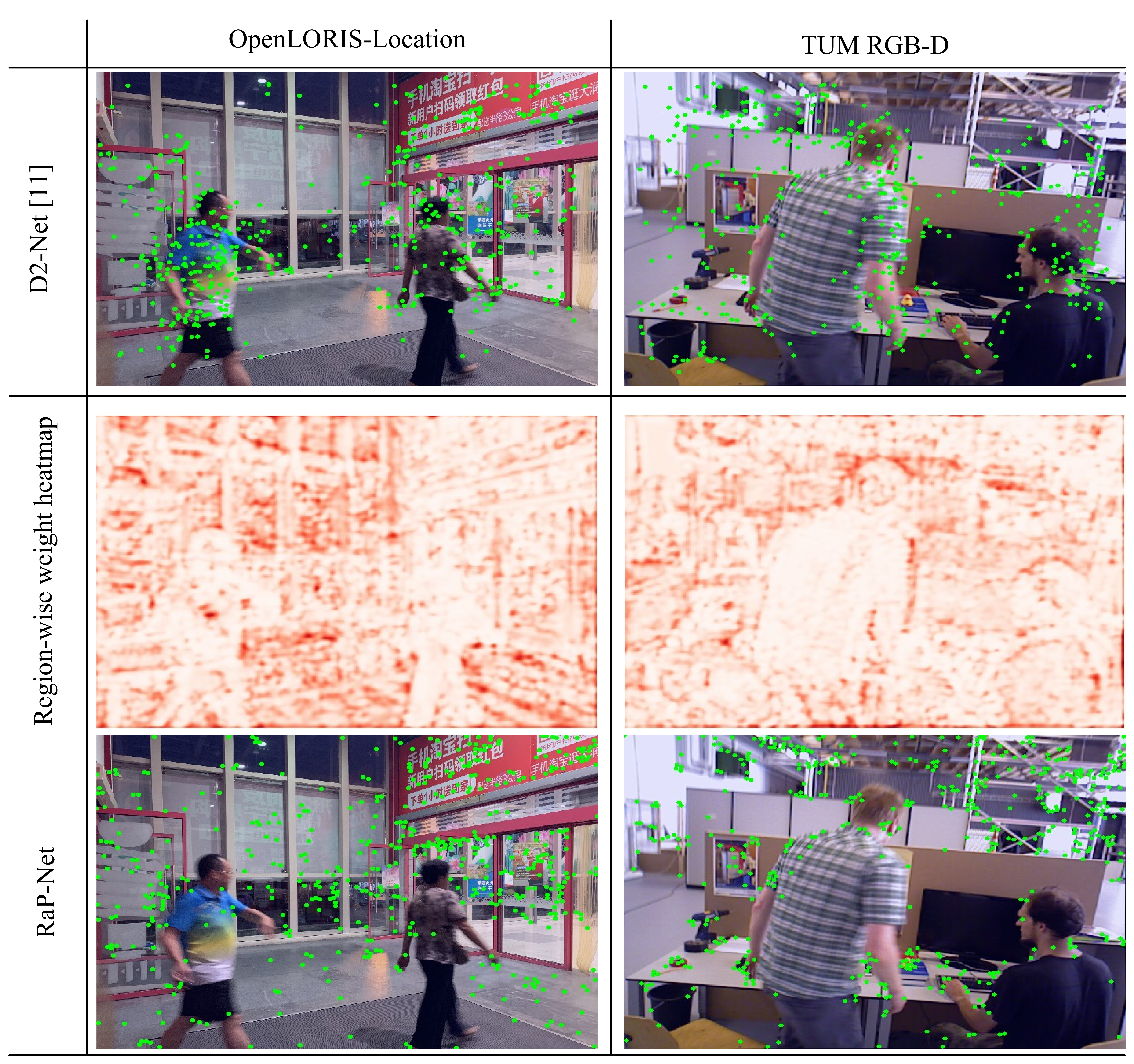}
\caption{Qualitative visualization between D2-Net and RaP-Net. Green dots are the extracted Top-500 features. The image of TUM RGB-D dataset \cite{sturm12iros} is not trained by RaP-Net.}
\label{fig:visualization}
\end{figure}

\begin{figure*}[t]
 \centering
 \includegraphics[width=0.98\textwidth]{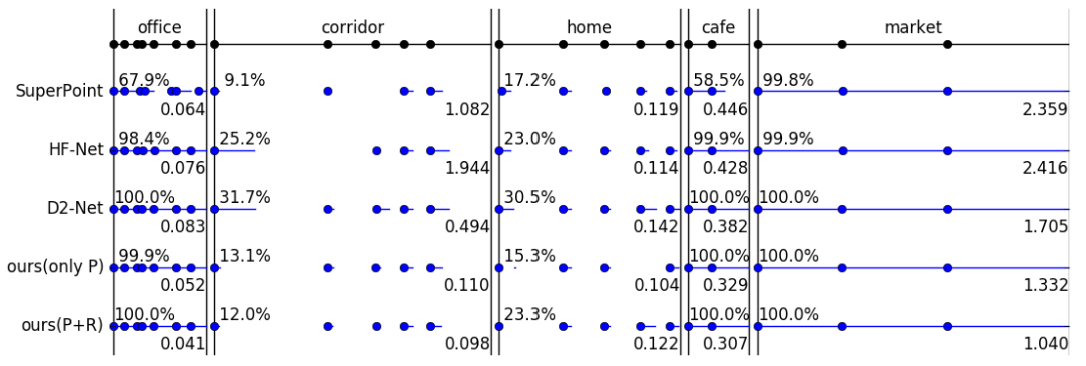}
 \caption{Per-sequence experimental results of visual SLAM (without re-localization and loop closure) with different features on the OpenLORIS-Scene dataset. Each black dot on the top line represents the start of one data sequence. For each algorithm, blue dots and lines indicate successful initialization and tracking. The percentage value on the top left of each scene is the average CR, larger means more robust. The float value on the bottom right is the average ATE RMSE, smaller means more accurate. The performances of fined-tuned SuperPoint is not shown here, because they fail to initialize and can not track in most of data sequences.}
 \label{fig:oloris-result}
 \end{figure*}

\begin{figure}[t!]
\centering
\includegraphics[width=0.48\textwidth]{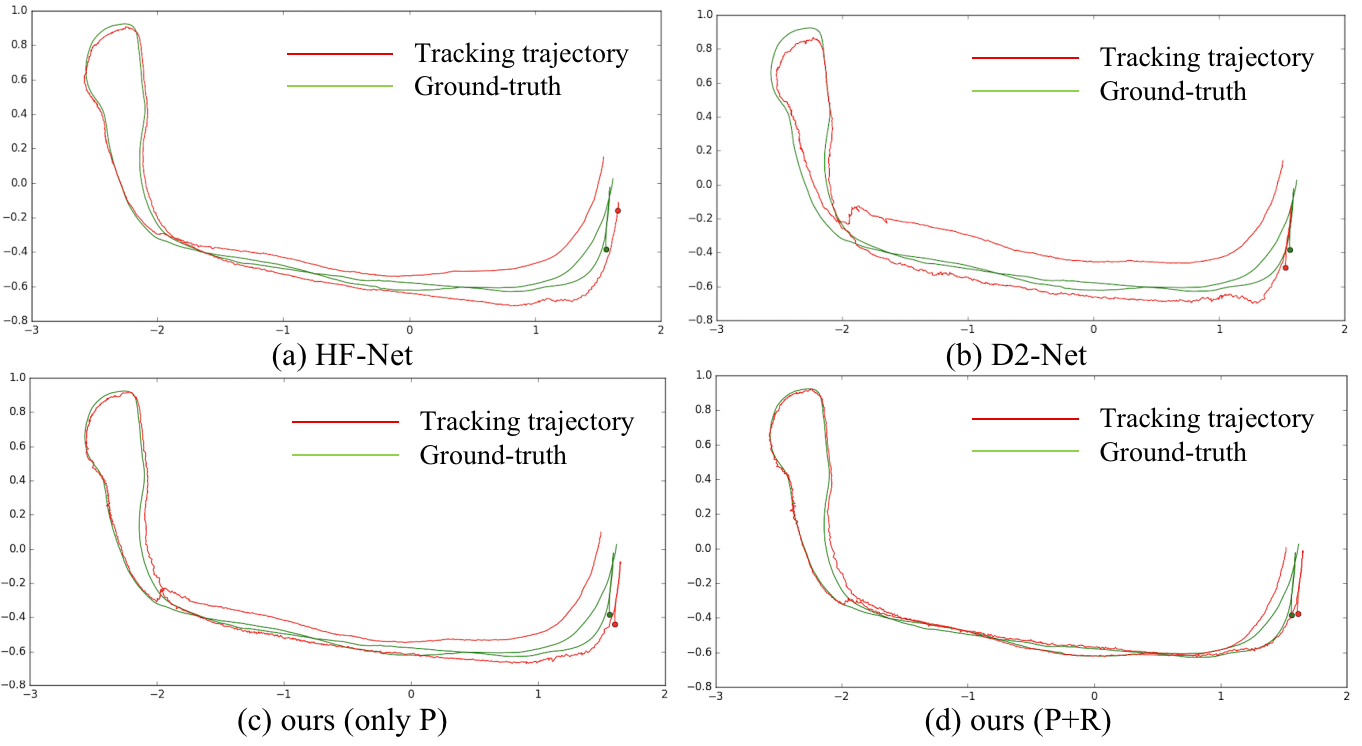}
\caption{Examples of tracking trajectory on office-1-5. The tracking performances of original and fine-tuned SuperPoint are not shown here, because they fail to initialize and can not track.}
\label{fig:office-1-5}
\end{figure}

\subsection{Indoor Localization}
The main motivation behind our work is to develop a local feature extraction approach to better handle indoor challenging conditions. Here, we provide some samples caught in indoor environments in Fig. \ref{fig:visualization}. Clearly, the region-wise weights in dynamic regions, e.g. pedestrians, are small than ones in invariable and static regions. Thus, the selected top-500 features are mainly located in appropriate regions for indoor localization.

For a more quantitative evaluation, we evaluate the performance of our proposed method in indoor visual localization, more specifically a visual odometry task. We compare it with state-of-the-art  CNN-based features, such as SuperPoint \cite{detone2018superpoint}, HF-Net \cite{sarlin2019coarse} and D2-Net \cite{dusmanu2019d2}. We use the publicly available model weights released by the original authors. As mentioned before, we finetuned SuperPoint on our proposed OpenLORIS-Location dataset since it is the only feature that can be trained without ground-truth transforms or correspondences between training images. We substitute the feature extraction module of ORB-SLAM2 with these methods and compare the localization performances on the OpenLORIS-Scene dataset (in a per-sequence evaluation fashion). Re-localization and loop closure detection modules are disabled to avoid the influence of global optimization. We use the Correct Rate (CR) and the RMSE of Absolute Trajectory Error (ATE RMSE) to evaluate the localization ability as in\cite{shi2020we}. We also tune the configuration of the SLAM system, including the number of features for initialization and the threshold of far point, to achieve the best localization performance for each feature extraction method. Especially, in order to get fair evaluation results, we remove the training images in the OpenLORIS-Scene dataset. By the way, these training samples just account for a very small proportion in the original dataset (almost below 0.5\% and maximal below 1\%). The results are shown in Fig. \ref{fig:oloris-result}.

In the OpenLORIS-Scene dataset, corridor and home contain a large proportion of textureless images, e.g. white walls, and it is impossible to track over the whole trajectory by only using visual features in SLAM. Thus, we focus on the results with the other scenes, i.e. office, cafe, and home. Those scenes contain many dynamic elements and frequent appearance changes. In the evaluation, the original and finetuned SuperPoint fail to track in most scenes due to too few extracted features. It concludes that simply adding indoor samples into the training set will not improve the performance of features in specific scenes. We need to apply a skillful mechanism to combine the excellent attribute of local features with task-specific attention, e.g. region-wise weight in our work. 

As in Fig. \ref{fig:oloris-result}, it is obvious that our proposed features outperform others from both aspects of robustness and accuracy of localization in these indoor scenes. The overall approach with region-wise weight (R+P) has the most accurate initial pose estimation and tracking performance over all compared algorithms. It surpasses the method that only used point-wise weight (P) since we consider not only the reliability of pixels, but also the invariability of larger regions. It concluded that our proposed region-wise weight can successfully handle various challenges in the indoor scene and lead to a better localization performance.

We also provide some examples of tracking trajectory on office-1-5 sequence in Fig. \ref{fig:office-1-5}. It can be clearly shown that our whole approach has the most accurate tracking trajectory in a typical indoor scene with dynamics and repetitive patterns, which indicates our innovation that using the region-wise invariability to re-weight point-wise reliability contributes to indoor localization. 

\section{Conclusions}
We have presented RaP-Net – a novel network for deep local feature extraction towards indoor visual localization. To overcome the interference from dynamic, repetitive, and texture-less regions in indoor pose estimation, the proposed approach extracts local features by rules of combining region-wise invariability and point-wise reliability, which extract features located in invariable and distinct regions. The proposed region-wise weight is the first attention mechanism successfully implemented in indoor localization to the best of our knowledge. To train such attention-based models in our proposed network, we introduce the OpenLORIS-Location dataset containing challenging elements for training attention mechanisms, such as dynamics, varying illuminations, and moderate viewpoint changes. The performance is evaluated on two typical tasks ranging from feature matching evaluation to visual odometry. Experimental results demonstrate significant improvements compared to state-of-the-art baselines in the localization system. In the future, we will improve the RaP-Net architecture to make it lighter and implement it on real robots to show its robustness and effectiveness thoroughly.

\bibliographystyle{IEEEtran}
\bibliography{references}

\end{document}